# Artistic Instance-Aware Image Filtering by Convolutional Neural Networks

Milad Tehrani, Mahnoosh Bagheri, Mahdi Ahmadi, Alireza Norouzi, Nader Karimi, Shadrokh Samavi

*Department of Electrical and Computer Engineering, Isfahan University of Technology, Isfahan, Iran 84156-83111*

*Abstract*— In the recent years, public use of artistic effects for editing and beautifying images has encouraged researchers to look for new approaches to this task. Most of the existing methods apply artistic effects to the whole image. Exploitation of neural network vision technologies like object detection and semantic segmentation could be a new viewpoint in this area. In this paper, we utilize an instance segmentation neural network to obtain a class mask for separately filtering the background and foreground of an image. We implement a top prior-mask selection to let us select an object class for filtering purpose. Different artistic effects are used in the filtering process to meet the requirements of a vast variety of users. Also, our method is flexible enough to allow the addition of new filters. We use pre-trained Mask R-CNN instance segmentation on the COCO dataset as the segmentation network. Experimental results on the use of different filters are performed. System's output results show that this novel approach can create satisfying artistic images with fast operation and simple interface.

## I. Introduction

Nowadays, people use smartphones as the main imaging device for capturing moments. The popularity of smartphones is due to their availability and simplicity of their use. Also, media-sharing social-network users are increasing rapidly. Every day people share millions of images on the applications like Instagram, Flicker, Pinterest, etc. For example, Instagram has had about 1 billion monthly active users in 2018, while their users were 800 million in September 2017, as declared by the company. Based on aesthetic desire in humans, most of the users tend to apply artistic effects to their photos. Bakhshi et al. [1] show that in social networks, filtered images have more potential to be viewed or get comments. Therefore, mentioned media-sharing applications contain various filters or effects to satisfy their users' requirements. Most of these effects are simple and work without any understanding of the image contents. But what if effects become complex or we want to change some features of images? Since the introduction of AlexNet by Krizhevsky et al. [2] in 2012, convolutional neural networks (CNNs) are used in many image processing tasks including object recognition and semantic segmentation. The revolution caused by CNNs, is so important that these days networks surpass human vision results in classification [3]. Many photo editing applications and sites, like Prisma [4] and Google Deep Dream [5], apply CNNs to create new complex effects by new approaches for images.

Along with works in creating new digital art effects, researches has done to answer why people want their photos to be filtered and have effects. The other question was what operations make photos more pleasant. Bakhshi et al. [1] examine the motivation of users' behind their applying of effects on images. Based on their results, most of the users use effects to make their viewers concentrate on some main parts of images by applying different effects. Authors in [6], show that digital media crafted by people are more pleasant to them. People love their edited photos because they are made by themselves, and those photos show their unique artistic viewpoint. Therefore, based on the above facts, new approaches to artistic effects should be considered. New methods should contain both aspects of simplicity and user-friendly interfaces.

A new approach is to exploit CNNs along with computer vision tasks, such as object detection, object classification, etc., to beautify daily photos.

In recent years, the vision community has introduced many technologies based on CNNS in different areas. One of the most challenging vision tasks is instance segmentation [7], which combines object detection and semantic segmentation. It renders vast information including class, number, location and inner pixels of the objects in the image. Mask R-CNN [8], introduced by the same authors of Faster R-CNN [9] as an extension of their initial work [9], which was intended for object instance segmentation. Ren et al. developed Faster R-CNN by using a small parallel Fully Convolutional Network (FCN) [10] path for creating segmentation masks. The mask generation path has a small computational cost, so the network is fast to use. As the result of the simplicity and speed of Mask R-CNN, it is utilized in many applications [11] [12].

In this paper, we exploit Mask R-CNN to generate unique class masks for user-specified filtering of foreground and background of an image. We introduce an object-selection method to choose the best object class for proper filtering. Our method is capable of creating new artistic photos through an object detection mechanism.

The remainder of this paper is organized as follows. In section 2 we describe our system structure and explain each block of the system precisely. Section 3 presents the system's qualitative results for some artistic processing. We conclude the paper in section 4.

## II. PROPOSED METHOD

In this section, we describe our system structure in a step-by-step manner. As Figure 1 shows, first, an input image is fed into the Mask R-CNN instance segmentation network [8]. Mask R-CNN provides system segmentation masks for every instance and labels each instance by its class identification number. Afterward, the system has a preprocessing step to provide the input of the artistic filtering stage. In the preprocessing stage, we build an object-selection block to choose which class should be filtered. In the next step, the selected-objects mask goes through a morphological processing block. Then the result of this stage and the original image are sent to the artistic filtering block. In the artistic filtering block, first, we extract the background and the foreground of the input image by using the object-selection mask. Then the user-specified background and foreground filters are applied to the extracted background and foreground regions. Finally, filtered foreground and background are fussed to build the final output.

### A. Instance Segmentation Network

Most of the instance segmentation methods, e.g. [13], are slow and not accurate enough. On the other hand, Mask R-CNN [8] introduced by Facebook AI team, is a simple and fast network. Since our system requires a fast segmentation process, along with classification, we utilize Mask R-CNN to detect and segment instances that exist in an input image. Mask R-CNN [8] has a simple structure that is shown in Figure 2.

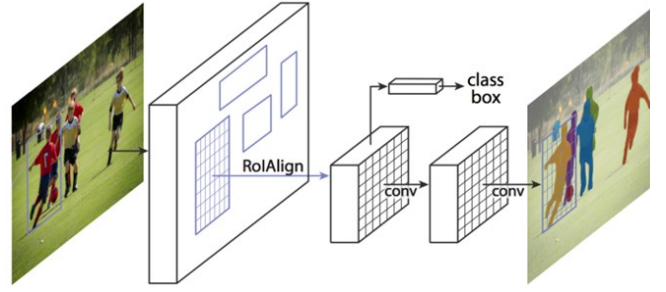

Fig. 2: Mask R-CNN Instance segmentation structure [8].

Mask R-CNN [8] generally comprises of 4 modules that are described in the following paragraphs.

*Backbone:* Mask R-CNN for feature extraction uses a standard CNN like ResNet101 [3] or ResNet50 [3]. Also, it uses the Feature Pyramid Network (FPN) [14] for improving its feature extraction process.

*Region Proposal Network (RPN):* A simple and light neural network which scans input image in a sliding-window manner and gives areas that contain objects. For making the system fast, the backbone feature map is fed into the RPN block. RPN uses about 200,000 overlapping region windows, called anchors, of different sizes and aspect ratios to scan the image and provides two outputs. The first output is the anchor class which is the foreground class that is likely to contain an object. The second output provides candidate objects' bounding boxes. These bounding boxes construct the system's region of interests (ROIs). The proposed method only picks up anchors with high probabilities of having objects. Then the algorithm refines their locations and sizes. Afterward, it uses non-max suppression for dealing with overlapping anchors, and the final proposals are passed into the next block.

*ROI Classification and Bounding Box Regression:* This stage gets ROIs given by the RPN and classifies the ROI's object to specific classes, such as a person, car, chair, etc. It also has a background class that the module discards. This stage also refines the size and location of bounding boxes to fit the object. Dividing a large resolution feature map into a smaller feature map using quantization causes the problem of misalignment on the boundaries due to the rounding operation. Mask R-CNN introduces ROIAlign instead of the ROIPool technique that is used in Faster R-CNN which improves the results.

*Segmentation Mask:* This block selects positive regions by the classifier and creates soft masks for them. In the training phase, the ground-truth masks are scaled down to 28×28 to calculate the loss, and during inferencing, the predicted masks are scaled up to the size of the ROI bounding box. After this stage, the final mask is ready for each instance.

### B. Preprocess

*Top Mask ID selection*

To choose which instances to be filtered we build an object-selection method, described in Figure 3. First, we create a unique class identification mask from the same class masks. The algorithm searches among input masks, and it composes masks with the same identifications (IDs). Then it calculates the unique mask areas and ranks them by size. We also prioritize classes based on the importance and number of the objects in the dataset [15]. For example, person class is the most prior one in our system. According to mask areas and priority, our system selects the top one and passes it to the next block.

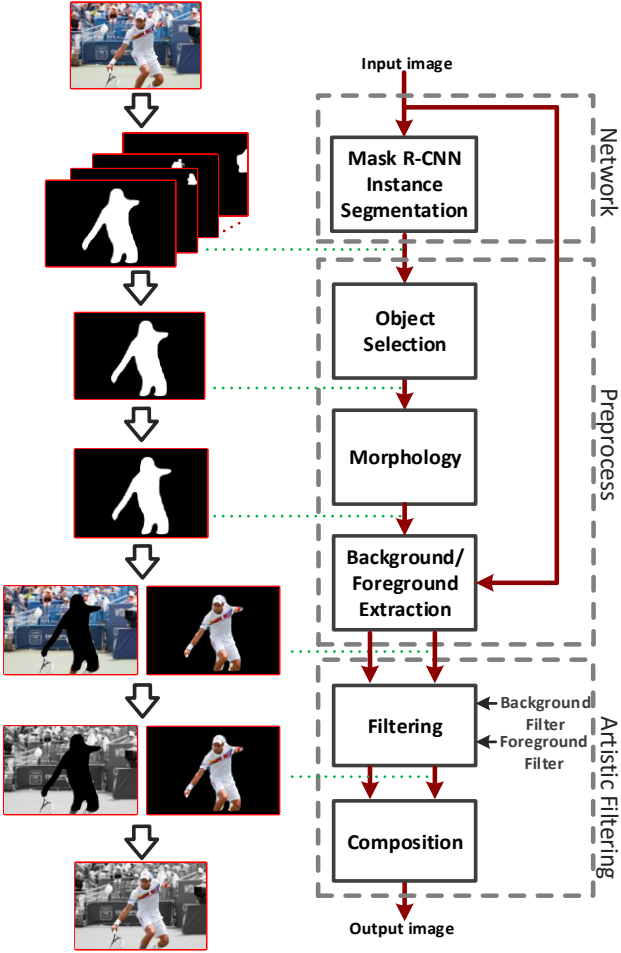

Fig. 1: The overall architecture of the proposed method.

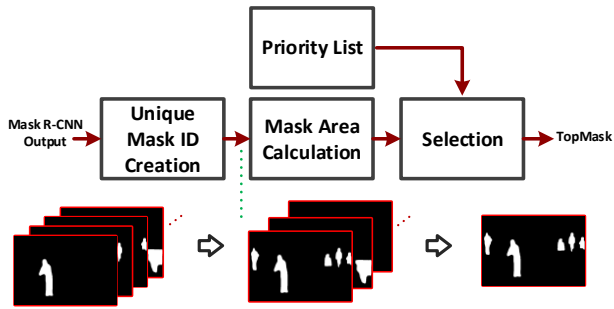

Fig. 3: Top mask selector diagram.

*Morphological processing*

We use morphological processing techniques including erosion, dilation, opening and closing on the object-selection mask to delete small isolated parts. Also, this process makes the output to become homogeneous and more appealing by smoothing the mask edges.

*Background and foreground extraction*

In this step, the system extracts background and foreground of the image, using the original image and the object-selection mask. This process is necessary for separate filtering of background and foreground.

### C. Artistic Filtering

The extracted background and foreground are fed into this stage along with the user-defined filters to get artistic effects. The user specifies background and foreground filters exclusively. We implemented a framework for the artistic filtering stage, so new filters can easily be added to the existing system. Some popular filters that are used in the current method are described in Table 1. After the filtering process, given the background and foreground filtered images are fused to form the system's output.

Table 1: Some filters contained in the system

| Filter keyword | Description |
|---|---|
| 'gray' | Convert image color-map to gray |
| 'bilateral' | A nonlinear noise reduction filter |
| 'edge-preserve' | Edge-preserve smoothing filter |
| 'median-blur' | A nonlinear denoising filter |
| 'gaussian-blur' | Blurring by a Gaussian function |
| 'detail-enhancement' | Enhancing details of the image |
| 'pencil-sketch' | Convert image to Pencil drawing like version |
| 'gray-blur' | Convert image color-map to gray and blur it by a Gaussian function |
| 'preserve' | Preserve image as exactly it is |

### III. EXPERIMENTAL RESULTS

We used ResNet101-FPN Mask R-CNN model backbone version [8], pre-trained on COCO dataset [15] which is a popular image dataset for object detection and segmentation. COCO has 80 object classes containing 330,000 images with 220,000 labeled ones. Consequently, it has a total of 1.5 million instances. We tested our method on a Core i5, 6G RAM Asus laptop with NVIDIA GeForce 740M GPU.

Because of the Mask R-CNN architecture, the proposed system can accept input images with different sizes. Hence its processing time may differ from image to image. Also, each filter requires a different operation time. As a result, the proposed method's execution time is variable for different images. We calculated the testing time for 100 images with different filters. The average execution time of 35.4s was achieved. Bypassing the morphological process stage would decrease this time to 33.7s.

Figure 4 illustrates the result of the proposed system with and without a morphological preprocessing stage. It shows that the masks produced by Mask R-CNN in some cases are a little inaccurate in the boundaries; hence erosion and dilation morphological techniques are needed to improve the results. For example Fig. 4. (d) shows an output result without morphology that is inaccurate in the bus borders, especially on the left and top edges.

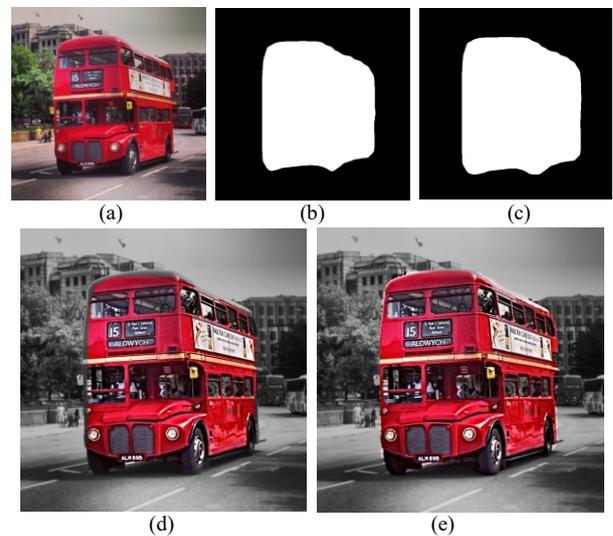

Fig. 4: Output results of the proposed method for generating a gray background and detail-enhancing of foreground, (a) input image (b) object-selection mask without morphology (c) mask with morphological process (d) output result without morphology (e) output result with morphology.

Some examples of our method results are shown in Fig. 5. These results created by using different foreground and background effects. As can be seen, results have good visual qualities and are accurately segmented by the Mask R-CNN and morphological techniques.

### IV. CONCLUSION

In recent years, application of photo-editing software packages has increased due to the vast use of social networks. Therefore, new approaches in this area are needed. In this paper, we introduced an instance segmentation based filtering method for adding a variety of artistic effects on images. We implemented a 'selection' structure to choose the most important class object for filtering. The artistic images created by our method, demonstrate high visual qualities which indicate that the proposed method is successful in generating artistic effects in images.

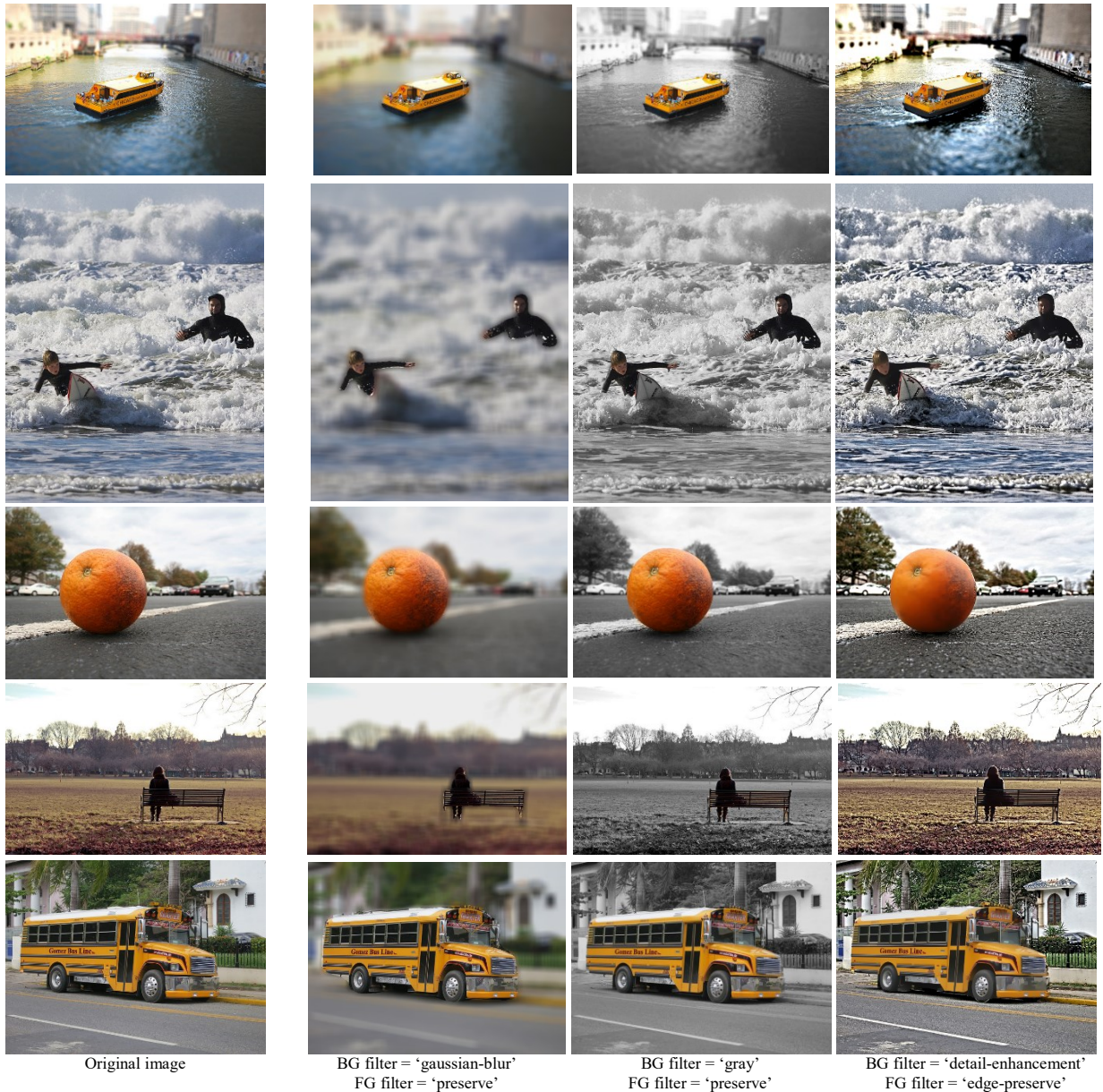

Fig. 5: Left column are input images ,other columns are outputs results with different foreground & background effects (BG & FG mean background & foreground)